\documentclass[utf8]{arxiv_submission} %

\usepackage[parskip=20pt,captionskip=0pt]{subfig}
\usepackage{url,lineno,microtype,subcaption}
\usepackage[onehalfspacing]{setspace}
\definecolor{myblue}{rgb}{0,0.08,0.5}
\usepackage[colorlinks = true,
            linkcolor = myblue,
            urlcolor  = myblue,
            citecolor = myblue]{hyperref}  
\usepackage[capitalize,noabbrev,nameinlink]{cleveref}            
\usepackage{booktabs}       %
\usepackage{amsfonts}       %
\usepackage{nicefrac}       %
\usepackage{microtype}      %
\usepackage{xcolor}         %
\usepackage{etoc}
\usepackage{fancyhdr,amsmath,graphicx} 
\usepackage{microtype}      %
\usepackage{colortbl}
\usepackage{multirow,makecell}
\usepackage{algpseudocode}
\usepackage{graphics}
\usepackage{epsfig}
\usepackage{xfrac}

\newcommand{\pubmed}{\textsf{Pubmed}\xspace}
\newcommand{\dblp}{\textsf{DBLP}\xspace}
\newcommand{\google}{\textsf{Google+}\xspace}
\newcommand{\fb}{\textsf{Facebook}\xspace}

\newcommand{\vgae}{\texttt{VGAE}\xspace}
\newcommand{\nodetovec}{\texttt{Node2vec}\xspace}
\newcommand{\fairpage}{\texttt{FairPR}\xspace}
\newcommand{\fairwalk}{\texttt{Fairwalk}\xspace}
\newcommand{\fairadj}{\texttt{FairAdj}\xspace}
\newcommand{\flip}{\texttt{FLIP}\xspace}
\newcommand{\fairegm}{\texttt{FairEGM}\xspace}
\newcommand{\method}{\texttt{FairLink}\xspace}
\newcommand{\ms}[2]{{#1\footnotesize{$\pm$#2}}}

\usepackage[ruled,vlined,linesnumbered]{algorithm2e}
\usepackage{color}
\usepackage{float}
\usepackage{wrapfig}

\usepackage{xspace}
\usepackage{bbm}

\definecolor{mygray}{gray}{0.9}
\newcolumntype{a}{>{\columncolor{mygray}}c}
\input{math_notation.sty}

\def\keyFont{\fontsize{8}{11}\helveticabold }
\def\firstAuthorLast{Yezi Liu} %
\def\Authors{Yezi Liu\,$^{1}$, Hanning Chen\,$^{1}$ and Mohsen Imani\,$^{1,*}$
}
\def\Address{
$^{1}$University of California, Irvine,  Irvine, CA, United States
}
\def\corrAuthor{Mohsen Imani}

\def\corrEmail{m.imani@uci.edu}

\begin{document}
\onecolumn
\firstpage{1}
\title[FairLink]{Promoting Fairness in Link Prediction with Graph Enhancement} 

\author[\firstAuthorLast ]{\Authors} %
\address{} %
\correspondance{} %

\extraAuth{}

\maketitle

\begin{abstract}
Link prediction is a crucial task in network analysis, but it has been shown to be prone to biased predictions, particularly when links are unfairly predicted between nodes from different sensitive groups. In this paper, we study the fair link prediction problem, which aims to ensure that the predicted link probability is independent of the sensitive attributes of the connected nodes. Existing methods typically incorporate debiasing techniques within graph embeddings to mitigate this issue. However, training on large real-world graphs is already challenging, and adding fairness constraints can further complicate the process. To overcome this challenge, we propose {\method}, a method that learns a fairness-enhanced graph to bypass the need for debiasing during the link predictor's training. {\method} maintains link prediction accuracy by ensuring that the enhanced graph follows a training trajectory similar to that of the original input graph. Meanwhile, it enhances fairness by minimizing the absolute difference in link probabilities between node pairs within the same sensitive group and those between node pairs from different sensitive groups. Our extensive experiments on multiple large-scale graphs demonstrate that {\method} not only promotes fairness but also often achieves link prediction accuracy comparable to baseline methods. Most importantly, the enhanced graph exhibits strong generalizability across different GNN architectures.

\tiny
 \keyFont{ \section{Keywords:} Fairness; Large-scale Graphs, Link Prediction.} 
\end{abstract}

\section{Introduction}
The scale of graph-structured data has expanded rapidly across various disciplines, including social networks~\citep{liben2003link}, citation networks~\citep{yang2016revisiting}, knowledge graphs~\citep{liu2023error,zhang2022contrastive}, and telecommunication networks~\citep{nanavati2006structural,xie2022explaining}. This growth has spurred the development of advanced computational techniques aimed at modeling, discovering, and extracting complex structural patterns hidden within large graph datasets. Consequently, research has increasingly focused on inference learning to identify potential connections, leading to the creation of algorithms that enhance the accuracy of link prediction~\citep{mara2020benchmarking,li2024evaluating}. Despite the strong performance of these models in link prediction, they can exhibit biases in their predictions~\citep{angwin2022machine,bose2019compositional}. These biases may result in harmful social impacts on historically disadvantaged and underserved communities, particularly in areas such as ranking~\citep{karimi2018homophily}, social perception~\citep{lee2019homophily}, and job promotion~\citep{clifton2019mathematical}. Given the widespread application of these models, it is crucial to address the fairness issues in link prediction.

Many existing studies have introduced the concept of fairness in link prediction and proposed algorithms to achieve it. For instance, {\fairadj}~\citep{li2021dyadic} introduces \textit{dyadic fairness}, which requires equal treatment in the prediction of links between two nodes from different sensitive groups, as well as between two nodes from the same sensitive group. These approaches are predominantly model-centric, incorporating debiasing methods during the training process~\citep{rahman2019fairwalk,masrour2020bursting,tsioutsiouliklis2021fairness,li2021dyadic,current2022fairegm}. However, promoting fairness in models trained on large-scale graphs is particularly challenging. State-of-the-art link predictors, often deep learning methods like GNNs, are already difficult to train on large graphs~\citep{zhang2021graphless,hu2021graph,ferludin2022tf,han2022mlpinit}. Introducing fairness considerations adds another layer of complexity, making the training process even more demanding. Therefore, model-centric approaches that attempt to enforce fairness during training may not be practical, as they introduce additional objectives that further complicate the already challenging training process~\citep{liu2023fairgraph}.

To address this challenge, we propose \method, a data-centric approach that incorporates dyadic fairness regularizer into the learning of the enhanced graph. This is achieved by optimizing a fairness loss function jointly with a utility loss. The utility loss is computed by evaluating the gradient distance~\citep{zhao2020dataset,jingraph,jin2022condensing}, which measures the differences in gradients between the enhanced and original graphs. This approach ensures that the task-specific performance is maintained in the learned graph~\citep{zhao2020dataset}. Additionally, the dyadic fairness loss directs the learning process towards generating a \textit{fair} graph for link prediction, while the utility loss ensures the preservation of link prediction performance. In contrast to model-centric approaches~\citep{zha2023data,jin2022empowering,zha2023data2}, which focus on designing fairness-aware link predictors, \method emphasizes the creation of a generalizable fair graph specifically for link prediction tasks.
We summarize our contributions as follows:
\begin{itemize}
\item This paper addresses the challenge of fair link prediction. While most existing methods concentrate on developing fairness-aware link predictors, we propose a novel data-centric approach. Our method focuses on constructing a fairness-enhanced graph. This graph can subsequently be used to train a link predictor without the need for debiasing techniques, while still ensuring fair link prediction.

\item To ensure fairness in the fairness-enhanced graph, {\method} optimizes a dyadic fairness loss function. Additionally, to preserve utility, {\method} minimizes the gradient distance between the fairness-enhanced graph and the original input graph. To improve the measurement of gradient distance, we introduce a novel scale-sensitive distance function.
\item The extensive experiments validate that, 1) the link prediction on the enhanced graph generated from {\method} is comparable with the link prediction on the input graph, 2) the fairness-utility trade-off of the enhanced graph is better than the baselines trained on the input graph, 3) the enhanced graph demonstrates strong generalizability, meaning it can achieve good fairness and utility performance on a test GNN architecture, even when it has been trained on a different GNN architecture.
\end{itemize}

\section{Preliminaries}
\begin{figure*}[!t]
\centering
\includegraphics[width=1.0\linewidth]{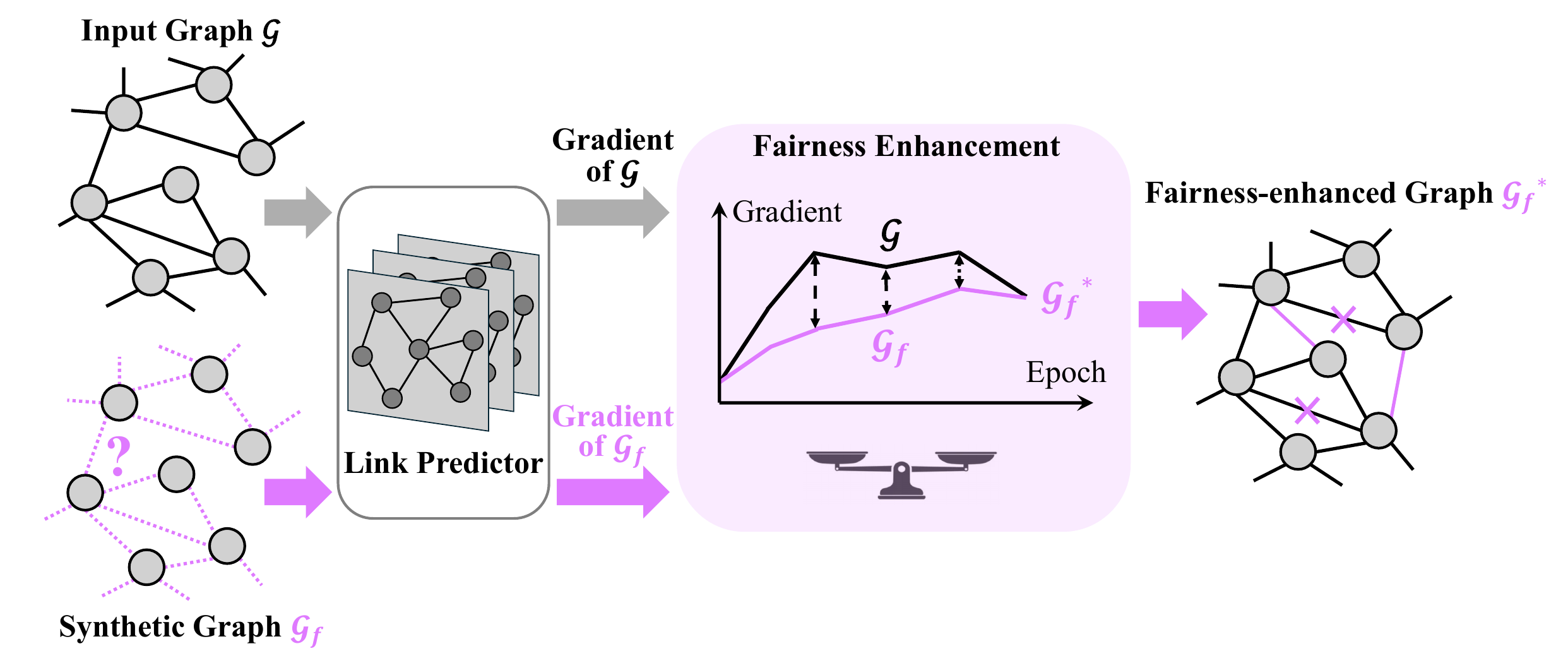}
\vspace{-2em}
\caption{The overall framework of {\method} aims to learn a fairness-enhanced graph in which both fairness is promoted and utility is preserved. Initially, a synthetic graph $\mathcal{G}_{f}$ is created with the same size as the input graph $\mathcal{G}$ and random link connections. Both the input graph and the synthetic graph are then fed into a trainable link predictor. The gradient of the cross-entropy loss with respect to the predictor's parameters is computed for both $\mathcal{G}$ and $\mathcal{G}_{f}$. The optimization of $\mathcal{G}_{f}$ involves minimizing a fairness loss in conjunction with the gradient distance between $\mathcal{G}$ and $\mathcal{G}_{f}$.
} \label{fig:framework}
\vspace{-1.5em}
\end{figure*}
In the following section, we will start by introducing the notations used in our study. Next, we will explore the concept of fairness within the context of link prediction, which involves estimating the probability of a connection between two nodes in a network. We will then extend the principles of fair machine learning to the fairness of link prediction.

\subsection{Notation} \label{sec:notation}
Let $\mathcal{G}=(\mathcal{V}, \mathcal{E}, {X})$ as a graph, where $\mathcal{V}$ is the set of $N$ nodes, $\mathcal{E} \subseteq \mathcal{V} \times \mathcal{V}$ is the edge set, ${X} \in \mathbb{R}^{N \times D}$ is the node features with $D$ dimensions.  ${A} \in\{0,1\}^{N \times N}$ is the adjacency matrix, where ${A}_{uv}=1$ if there is an edge between nodes $u$ and $v$. $(u, v)$ denotes an edge between node $u$ and node $v$. ${S}\in\mathbb{R}^{N \times K}$ is the vector containing sensitive attributes, $K$ is the number of sensitive attributes can take on, (e.g., $S_u\in\{\text{Female, Male, Unkown}\}$ for node $u$). $g(\cdot,\cdot): \mathbb{R}^H \times \mathbb{R}^H \rightarrow \mathbb{R}$ is the bivariate link predictor, and $g(z_u,z_v)$ is the predicted probability of an edge $(u, v)\in\mathcal{E}$ in a given graph, where $z_u$ and $z_v$ are the node embedding vectors with dimension $H$ for node $u$ and $v$. The \textit{problem} of fair link prediction aims to learn a synthetic graph $\mathcal{G}_{f}=(\mathcal{V}_f, \mathcal{E}_f, {X}_f)$, where a link predictor $g(\cdot,\cdot)$ trained on $\mathcal{G}_f$ will obtain comparable performance with it trained on the original graph $\mathcal{G}$, and the link predictions are fair. In our experiemnts, $|\mathcal{V}_f|=|\mathcal{V}|$ and ${X}_f \in \mathbb{R}^{N \times D}$.

\subsection{Fairness in Link Prediction} \label{sec:fair_metric}
Previous research in fair machine learning has typically defined fairness in the context of binary classification as the condition where the predicted label is independent of the sensitive attribute. In the domain of link prediction, which involves estimating the probability of a link between pairs of nodes in a graph, fairness can be extended by ensuring that the estimated probability is independent of the sensitive attributes of the two nodes involved. In this subsection, we introduce two fairness concepts relevant to link prediction: demographic parity and equal opportunity.
\subsubsection{Demographic Parity}
Demographic Parity (DP) requires that predictions are independent of the sensitive attribute. It has been extensively applied in previous fair machine learning studies, and by replacing the classification probability with link prediction probability, it can be simply extended in the context of link prediction. It is also named \textit{dyadic fairness} in previous literature~\citep{li2020dyadic}. In the context of link prediction, the DP requires the prediction probabilities should be independent of the attributes of both nodes in the link. 
\begin{equation}
    P(g(u,v)|S_u=S_v)=P(g(u,v)|S_u\neq S_v).
    \label{eq:dyadic}
\end{equation}
Ideally, achieving dyadic fairness entails predicting intra- and inter-link relationships at the same rate from a set of candidate links. The metric used to assess dyadic fairness in link prediction is as follows:
\begin{equation}
\Delta_{\mathit{DP}}=|P(g(u, v) \mid S_u=S_v)-P(g(u, v) \mid S_u\neq S_v)|.
\end{equation}
\subsubsection{Equal Opportunity}
Equal Opportunity (EO) ensures that the probability of a positive outcome is consistent across different sensitive groups, given the actual links in the graph. This approach aims to prevent any group from being unfairly disadvantaged. The formal definition of EO in link prediction is as follows:
\begin{equation}
   P(g(u, v) \mid S_u=S_v)=P(g(u, v) \mid S_u\neq S_v, (u,v)\in\mathcal{E}).
    \label{eq:eo}
\end{equation}
Specifically, EO requires that the predicted probability $g(u,v)$ of a true link $(u,v)\in\mathcal{E}$ should be consistent across different sensitive groups. The method for assessing distance of EO fairness in link prediction is defined as follows:
\begin{equation}
\Delta_{\mathit{EO}}=|P(g(u, v) \mid S_u=S_v)-P(g(u, v) \mid S_u\neq S_v), (u,v)\in\mathcal{E}|.
\end{equation}

\section{Fairness-enhanced Graph Learning}
\begin{wrapfigure}[13]{r}{0.32\textwidth}
     \vspace{-2em}
	\centering
    \includegraphics[width=0.28\textwidth]{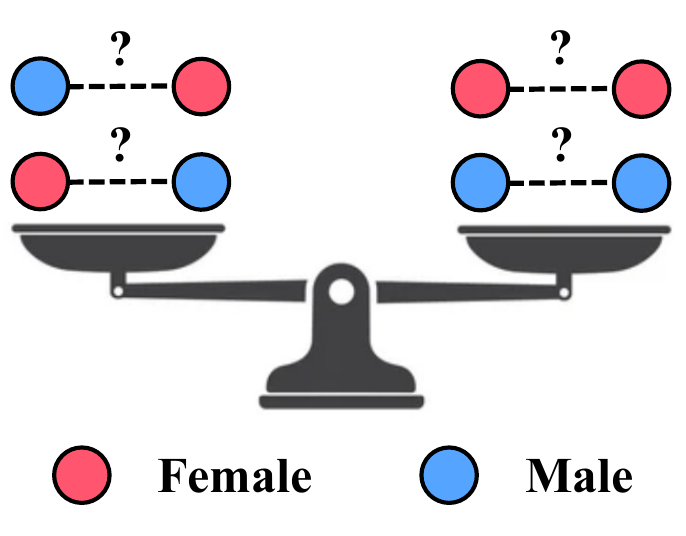}
        \vspace{-1em}
	\caption{Fair link prediction objective in {\method}: Ensure equal probability for links between nodes from different sensitive groups and those from the same group.} \label{fig:task}
\end{wrapfigure}
In this section, we provide a comprehensive description of {\method}. Our objectives are twofold: (1) ensuring fairness within the fairness-enhanced graph and (2) preserving the utility of the fairness-enhanced graph. Specifically, our approach involves constructing a fairness-enhanced graph from the input graph to improve fairness in link prediction. To achieve the first objective, {\method} incorporates a dyadic regularization term that promotes fairness. For the second objective, {\method} maintains utility by minimizing the gradient distance between the input graph and the enhanced graph. Additionally, we introduce a novel scale-sensitive distance function to optimize the learned graph and measure the gradient distance effectively. To simplify the notation, we omit the training epoch $t$ when introducing the loss function at a specific epoch. A framework of {\method} is provided in~\cref{fig:framework}.
\subsection{Fairness Enhancement}
In this subsection, we describe how to equip the learned graph with fairness-aware properties. This is achieved by incorporating a dyadic fairness regularizer, as specified in \cref{eq:dyadic}, into the learning process of the fairness-enhanced graph. Further details on this process can be found in \cref{sec:fair_metric}. The schematic diagram in \cref{fig:task} illustrates the fairness objective of the fairness-enhanced graph learning within {\method}.

The concept of fairness constraint has been investigated in~\citep{zafar2015fairness,zafar2017fairness} by minimizing the disparity in fairness between users' sensitive attributes and the signed distance from the users' features to the decision boundary in fair linear classifiers. In this paper, we incorporate a fairness regularizer derived from $\Delta_{\mathit{DP}}$~\citep{chuang2021fair,zemel2013learning}, which quantifies the difference in the average predictive probability between various demographic groups. The fairness loss function $\mathcal{L}_{\mathit{fair}}$ at training epoch $t$ is defined as follows:
\begin{equation}\label{equ:dp_pro}
    \begin{split}
        \mathcal{L}_{\mathit{fair}} = |\mathbb{E}_{u, v \sim \mathcal{V} \times \mathcal{V}}[g(u,v)|s_u=s_v]-\mathbb{E}_{u, v \sim \mathcal{V} \times \mathcal{V}}[g(u,v)|s_u \neq s_v]|,
    \end{split}
\end{equation}
where $\mathbb{E}$ is estimated $\mathbb{E}_{u, v \sim \mathcal{V} \times \mathcal{V}}$ is generated from a link between any node pair in the graph.
where $\hat{\mathbf{Y}}$ represents the prediction probability of the downstream task. The variable $N$ denotes the total number of instances, while $N_{\mathbf{s}=0/1}$ refers to the total number of samples in the group associated with the sensitive attribute values of $0/1$ respectively. The fundamental requirement for $\Delta_{\mathit{DP}}$ is that the average predictive probability $\hat{\mathbf{Y}}$ within the same sensitive attribute group serves as a reliable approximation of the true conditional probability $\mathrm{P}(\hat{\mathbf{Y}}=1|\mathbf{S}=0)$ or $\mathrm{P}(\hat{\mathbf{Y}}=1|\mathbf{S}=1)$.

\subsection{Utility Preserving}
In this section, we address the \textit{first objective}: determining how to learn a fairness-enhanced graph such that a link predictor trained on it exhibits comparable performance to one trained on the input graph. 
{\method} first computes the link prediction loss for the original graph, denoted as $\mathcal{L}(\mathcal{G})$, by calculating the cross-entropy loss between the predicted link distribution (based on the dot product scores of the node embeddings) and the actual link distribution. Similarly, the link prediction loss for the synthetic graph, $\mathcal{L}(\mathcal{G}_f)$, is computed in the same manner. The gradients of both graphs with respect to the link predictors' weights, denoted as $\nabla_{\theta}\mathcal{L}(\mathcal{G})$ and $\nabla_{\theta}\mathcal{L}(\mathcal{G}_f)$, are then obtained. We define the utility loss $\mathcal{L}_{\mathit{util}}$ as the sum of the distances between these gradients across all training epochs.

Previous studies have utilized Cosine Distance to measure the distance between two gradients~\citep{zhao2020dataset,jingraph,liu2024tinygraph,shentinydata}. While effective, Cosine Distance is scale-insensitive, meaning it ignores the magnitude of the vectors. Since the magnitude of the gradient is critical for optimization, incorporating it into the distance measurement is important. To address this limitation, we propose a combined approach that integrates Cosine Distance with Euclidean Distance, which accounts for vector magnitudes. Thus, the revised distance function \( D \) is defined as:
$$D\left(\nabla_{{\theta_{t}}}\mathcal{L}(\mathcal{G}), 
\nabla_{{\theta_{t}}}\mathcal{L}(\mathcal{G}_{f})\right)=D_{\mathit{cos}} +\gamma D_{\mathit{euc}},$$
where $D_{\text{cos}}$ denotes the Cosine Distance, $D_{\text{euc}}$ denotes the Euclidean Distance, $\gamma$ serves as a trade-off hyperparameter, and $\theta_{t}$ is the trainable parameters for link predictor at training epoch $t$. The definitions of these distances are as follows:
\begin{equation}
\begin{split}
D_{\mathit{cos}}\left(\nabla_{\theta_t} \mathcal{L}(\mathcal{G}), \nabla_{\theta_t} \mathcal{L}(\mathcal{G}_{f})\right)&=\sum_i\left(1-\frac{\nabla_{\theta_t} \mathcal{L}(\mathcal{G})_{i} \cdot \nabla_{\theta_t} \mathcal{L}(\mathcal{G}_{f})_i}{\left\|\nabla_{\theta_t} \mathcal{L}(\mathcal{G})_i\right\|\left\|\nabla_{\theta_t} \mathcal{L}(\mathcal{G}_{f})_i\right\|}\right),\\
D_{\mathit{euc}}\left(\nabla_{\theta_t} \mathcal{L}(\mathcal{G}), \nabla_{\theta_t} \mathcal{L}(\mathcal{G}_{f})\right)&=
\left\| \nabla_{\theta_t} \mathcal{L}(\mathcal{G})_i-\nabla_{\theta_t} \mathcal{L}(\mathcal{G}_{f})_i \right\|.
\end{split}
\end{equation}

The utility loss at a specific epoch $t$, denoted as $\mathcal{L}_{\text{util}}$, is computed by summing the gradient distances between $\mathcal{G}$ and $\mathcal{G}_{f}$ across all training epochs. It is formally defined as follows:
$$
\mathcal{L}_{\mathit{util}}= D\left(\nabla_{\theta_t} \mathcal{L}(\mathcal{G}), \nabla_{\theta_t} \mathcal{L}\left(\mathcal{G}_{f}\right)\right).
$$
Minimizing the utility loss ensures that the training trajectory of $\mathcal{G}_{f}$ closely follows that of $\mathcal{G}$, leading to parameters learned on $\mathcal{G}_{f}$ closely approximating those learned on $\mathcal{G}$. As a result, $\mathcal{G}_{f}$ preserves the essential information of the input graph $\mathcal{G}$.
\subsection{Optimization}
Optimizing a fairness-enhanced graph directly is computationally expensive due to the quadratic complexity involved in learning ${\bf A}_{f}$. To address this challenge, previous work~\citep{jingraph} proposed modeling ${\bf A}_{f}$ as a function of ${\bf X}_{f}$. We further simplify this approach by using a multi-layer perceptron parameterized by $\psi$ with a sigmoid activation function to model the relationship, thereby reducing the computational burden. Thus, the final loss function is as follows:
\begin{equation}
   \min_{\mathbf{X}_{f}, \psi}\mathbb{E}_{\theta_{0} \sim P_{\theta_{0}}}\left[\sum_{t=0}^{T-1} \left(\mathcal{L}_{\mathit{util}} + \alpha\mathcal{L}_{\mathit{fair}} + \beta \|\theta_{t}\|^2\right)\right],\label{eq:final_loss}
\end{equation}
where $T$ is the total training epochs, $\alpha$ and $\beta$ are hyperparameters that govern the influence of two critical aspects: the gradient matching loss and the $L_2$ norm regularization, respectively.

Jointly optimizing ${\bf X}_{f}$ and $\psi$ is often challenging due to the interdependence between them. To overcome this, we employ an alternating optimization strategy. We first update ${\bf X}_{f}$ for $\tau_1$ epochs, then update $\psi$ for $\tau_2$ epochs. This process is repeated alternately until the stopping criterion is satisfied.

\subsection{Fair Link Prediction}
To achieve fair link prediction, we first use the fairness-enhanced graph  $\mathcal{G}_{f}$ to train a link predictor. This link predictor can differ in architecture from the model that produced $\mathcal{G}_{f}$ and does not necessarily incorporate fairness considerations. In this paper, we define the link prediction function $g(\cdot, \cdot)$ as the inner product between the embeddings of two nodes $u$ and $v$, for each node pair $(u, v) \in \mathcal{V} \times \mathcal{V}$. Specifically, the function is defined as $g(u, v) = u^{\top} \Sigma v$, where $\Sigma$ is a positive-definite matrix that scales the input vectors directionally. In our implementation, $\Sigma$ is set to an identity matrix, simplifying $g(\cdot, \cdot)$ to the dot product, which is commonly used in link prediction research~\citep{trouillon2016complex, kipf2016variational}.

\noindent\textbf{Dicussion.}
In this paper, the fairness-enhanced graph produced by {\method} retains the same size as the input graph, as discussed in~\cref{sec:notation}. To facilitate fairness-aware training on large-scale graphs, our approach concentrates on learning a fairness-enhanced graph that can be reused, thereby eliminating the need for repeated debiasing in future training with different link predictors. Future work could investigate methods for learning a smaller, fairness-enhanced graph derived from large-scale real-world graphs.

\section{Experiments}
In this section, we evaluate the effectiveness of {\method} on four large-scale real-world graphs. We focus on assessing its performance in link prediction and fairness, as well as the trade-off between fairness and utility by comparing {\method} with seven baseline methods. Additionally, we examine the generalizability of the graphs generated by {\method} by applying them to various GNN architectures.
\subsection{Experimental Setup}
\noindent\textbf{Datasets.}
We consider four large-scale graphs that have been extensively used in previous studies on fair link prediction. These graphs span a diverse range of domains, including citation networks, co-authorship networks, and social networks, each characterized by different sensitive attributes. We consider the nodes that take minority as the protected node group (e.g., Female nodes {\google} and male nodes in the {\fb}). The statistics of the datasets are in~\cref{tab:main_exp}. 
\begin{itemize}
    \item \textbf{{\pubmed}}~\footnote{{\pubmed}: \url{https://linqs.org/datasets/}}: {\pubmed} is a dataset where each node represents an article, characterized by a bag-of-words feature vector. An edge between two nodes indicates a citation between the corresponding articles, regardless of direction. The topic of an article, i.e., its category, is used as the sensitive attribute in this dataset.

    \item \textbf{{\dblp}}~\footnote{{\dblp}: \url{https://dblp.dagstuhl.de/xml/}}~\citep{tang2008arnetminer}: {\dblp} is a co-authorship network constructed from the DBLP computer science bibliography database. The network comprises nodes representing authors extracted from papers accepted at eight different conferences. An edge exists between two nodes if the corresponding authors have collaborated on at least one paper. The sensitive attribute in this dataset is the continent of the author's institution. 

    \item \textbf{{\google}}~\footnote{{\google}: \url{https://snap.stanford.edu/data/ego-Gplus.html}}~\citep{leskovec2012learning}: {\google} is a social network dataset. The data was collected from users who chose to share their social circles, where they manually categorized their friends on the Google+ platform.

    \item \textbf{{\fb}}~\footnote{{\fb}: \url{https://snap.stanford.edu/data/ego-Facebook.html}}~\citep{leskovec2012learning}: {\fb} is a dataset that contains anonymized feature vectors for each node, representing various attributes of a person’s Facebook profile.
\end{itemize}

\noindent\textbf{Baselines.}
We compare with two link prediction approaches, {\vgae} and {\nodetovec}, and five state-of-the-art fair link prediction methods, {\fairpage}, {\fairwalk}, {\fairadj}, {\flip}, and {\fairegm}. 
    
\begin{itemize}
\item \textbf{Link prediction methods}: We consider two popular link prediction baselines:1) The Variational Graph Autoencoder (\vgae)~\citep{kipf2016variational}, which is based on the variational autoencoder model. \vgae uses a GNN as the inference model and employs latent variables to reconstruct graph connections. 2) \nodetovec~\citep{grover2016node2vec}, a widely-used graph embedding approach based on random walks. It represents nodes as low-dimensional vectors that capture proximity information, enabling link prediction through these node embeddings.

    \item \textbf{Fair link prediction methods}: To evaluate fairness in link prediction, we compare against five state-of-the-art approaches:
    1) {\fairpage}~\citep{tsioutsiouliklis2021fairness}, which extends the PageRank algorithm by incorporating group fairness considerations. 
    2) \fairwalk~\citep{rahman2019fairwalk}, built upon {\nodetovec}, modifies transition probabilities during random walks based on the sensitive attributes of a node's neighbors to promote fairness. 
    3) {\fairadj}~\citep{li2021dyadic}, a regularization-based link prediction algorithm, ensures dyadic fairness by maintaining the utility of link prediction through a {\vgae}, while enforcing fairness with a dyadic loss regularizer. 
    4) {\flip}~\citep{masrour2020bursting} enhances structural fairness in graphs by reducing homophily, and evaluates fairness by measuring reductions in modularity. 
    5) {\fairegm}~\citep{current2022fairegm}, a collection of three methods, emulates different types of graph modifications to improve fairness. In our experiments, we use Constrained Fairness Optimization (GFO) as the representative method from this collection.
\end{itemize}
For a detailed discussion of the fair link prediction baselines, please refer to~\cref{sec:related_fairlink}.

\noindent\textbf{Metrics.} 
To evaluate the accuracy of link prediction, we use two metrics: the F1-score and the area under the receiver operating characteristic curve (AUC)~\citep{current2022fairegm,li2021dyadic,masrour2020bursting}. For assessing group fairness, we adopt two additional metrics: the difference in demographic parity ($\Delta_{\mathit{DP}}$)~\citep{feldman2015certifying} and the difference in equal opportunity ($\Delta_{\mathit{EO}}$)~\citep{hardt2016equality}, as introduced in~\cref{sec:fair_metric}. Lower values of $\Delta_{\mathit{DP}}$ and $\Delta_{\mathit{EO}}$ indicate better fairness, making these metrics crucial for evaluating the fairness of the model.
\begin{table*}[t]
    \begin{minipage}{\columnwidth}
        \centering
        \caption{Link prediction and fairness results on large-scale graphs. An upward arrow ($\uparrow$) indicates that a higher value is better, while a downward arrow ($\downarrow$) signifies the opposite. For each metric, the best results are highlighted in \textbf{bold}, and the runner-up results are \underline{underlined}.}
        \label{tab:main_exp}
        \setlength{\tabcolsep}{2pt}
        \scalebox{0.98}
{
\begin{tabular}{r|cc|ccccc|c}
\toprule
\textbf{Metric}                        &{\vgae}                       &  {\nodetovec}      & {\fairpage}          &  {\fairwalk}      & {\fairadj}                   & {\flip}                     & {\fairegm}        & \textbf{\method}        \\
\midrule
\multicolumn{9}{c}{{{\pubmed}} \  \#Nodes: $19,717$  \ \#Edges: $88,648$  \ Sensitive Attribute: Topic }                                                                                                                                            \\
\midrule
F1 ($\uparrow$)                        & \textbf{\ms{93.18}{1.07}} & \ms{86.50}{1.48}  & \ms{83.33}{2.79}      & \ms{85.20}{2.53}  & \ms{84.25}{1.21}             & \ms{83.48}{1.79}            & \ms{83.70}{1.68}     &\underline{\ms{90.46}{1.67}}       \\
AUC ($\uparrow$)                       & \textbf{\ms{96.20}{0.85}}    & \ms{93.27}{1.23}  & \ms{88.21}{0.62}      & \ms{91.43}{1.11}  & \ms{90.53}{1.03}             & \ms{87.44}{1.36}            & \ms{88.12}{2.33}  & \underline{\ms{95.24}{1.65}}    \\
$\Delta_{\mathit{DP}}$ ($\downarrow$)  & \ms{20.88}{12.48}            & \ms{19.14}{11.93} & \ms{17.31}{6.32}      & \ms{18.42}{8.65}  & \underline{\ms{14.73}{5.98}} & \ms{15.42}{7.69}            & \ms{17.52}{6.30}  & \textbf{\ms{5.42}{2.65}}         \\
$\Delta_{\mathit{EO}}$ ($\downarrow$)  & \ms{18.84}{10.98}            & \ms{20.33}{8.74}  & \ms{15.39}{9.52}      & \ms{20.18}{7.75}  & \underline{\ms{16.39}{4.64}} & \ms{19.43}{8.01}            & \ms{19.29}{9.44}  & \textbf{\textbf{\ms{4.86}{1.34}}}                  \\
\midrule
\multicolumn{9}{c}{{\dblp} \ \#Nodes: $13,015$  \ \#Edges: $79,972$  \ Sensitive Attribute: Continent }                                                                                                                                            \\
\midrule
F1 ($\uparrow$)                        & \textbf{\ms{82.23}{1.66}}    & \ms{78.15}{1.72}  & \ms{80.05}{1.27}      & \ms{80.88}{2.81}  &\ms{81.62}{1.58}  & \ms{77.62}{1.71}     & \ms{80.45}{0.92}  &\underline{\ms{81.69}{1.55}}       \\
AUC ($\uparrow$)                       & \textbf{\ms{90.77}{1.82}}    & \ms{83.21}{2.94}  & \ms{72.43}{1.30}      & \ms{88.39}{1.59}  & \ms{84.51}{2.25} & \ms{78.14}{3.41}     & \ms{80.43}{2.62} &\underline{\ms{88.72}{1.76}}      \\
$\Delta_{\mathit{DP}}$ ($\downarrow$)  & \ms{7.42}{3.95}              & \ms{8.43}{5.25}   & \ms{11.65}{4.33}      & \ms{9.86}{4.04}   & \underline{\ms{3.55}{3.37}}  & \ms{6.34}{4.22}             & \ms{5.82}{5.33}   & \textbf{\ms{1.32}{0.45}}        \\
$\Delta_{\mathit{EO}}$ ($\downarrow$)  & \ms{8.53}{3.60}              & \ms{7.22}{4.37}   & \ms{9.37}{5.24}       & \ms{7.10}{3.57}   & \ms{5.82}{3.91}              & \underline{\ms{5.39}{4.37}} & \ms{7.33}{6.32}   & \textbf{\ms{2.19}{1.01}}      \\
\midrule
\multicolumn{9}{c}{{\google}  \ \#Nodes: $4,938$  \ \#Edges: $547,923$  \ Sensitive Attribute: Gender}                                                                                                                                           \\
\midrule
F1 ($\uparrow$)                        &\textbf{\ms{88.33}{1.21}}     & \ms{81.11}{1.50}  & \ms{76.22}{1.36}      & \ms{82.47}{1.08}  & \ms{84.77}{1.19}             & \ms{78.35}{2.02}            & \ms{80.69}{1.53}  & \underline{\ms{85.34}{0.81}}      \\
AUC ($\uparrow$)                       & \textbf{\ms{94.85}{0.91}}    & \ms{88.74}{2.84}  & \ms{67.29}{1.53}      & \ms{93.01}{0.58}  &\ms{93.37}{0.22}              & \ms{81.86}{1.54}            & \ms{80.26}{1.61}  & \underline{\ms{94.42}{1.86}}    \\
$\Delta_{\mathit{DP}}$ ($\downarrow$)  & \ms{6.42}{2.05}              & \ms{7.88}{4.72}   & \ms{7.14}{1.83}       & \ms{5.61}{4.20}   & \ms{3.79}{1.22}              & \textbf{\ms{1.19}{1.93}}    & \ms{4.55}{2.11}   & \underline{\ms{1.42}{0.96}}      \\
$\Delta_{\mathit{EO}}$ ($\downarrow$)  & \ms{7.92}{4.48}              & \ms{9.35}{3.19}   & \ms{6.35}{3.09}       & \ms{4.42}{1.93}   & \ms{3.76}{1.47}              & \underline{\ms{2.21}{1.12}} & \ms{5.37}{3.65}   & \textbf{\ms{1.01}{0.75}}       \\
\midrule
\multicolumn{9}{c}{{\fb} \ \#Nodes: $1,045$  \ \#Edges: $53,498$ \ Sensitive Attribute: Gender} \\ \midrule
F1 ($\uparrow$)                        & \textbf{\ms{82.41}{1.23}}    & \ms{79.35}{0.95}  & \ms{76.22}{1.30}      & \ms{78.11}{0.78}  & \ms{81.14}{1.23}             & \ms{78.5}{1.42}             & \ms{79.77}{2.92}   & \underline{\ms{82.37}{0.41}}         \\
AUC ($\uparrow$)                       & \textbf{\ms{94.66}{0.55}}    & \ms{90.57}{1.24}  & \ms{70.30}{1.09}      & \ms{91.56}{0.63}  & \ms{92.53}{1.49}             & \ms{83.0}{1.51}             & \ms{85.42}{1.45}   & \underline{\ms{93.73}{1.72}}    \\
$\Delta_{\mathit{DP}}$ ($\downarrow$)  & \ms{2.03}{0.81}              & \ms{1.70}{1.43}   & \ms{2.33}{1.91}       & \ms{1.97}{1.51}   & \ms{1.77}{0.81}              & \underline{\ms{1.17}{0.55}} & \ms{2.21}{1.05}    & \textbf{\ms{0.83}{0.36}}       \\
$\Delta_{\mathit{EO}}$ ($\downarrow$)  & \ms{3.78}{2.15}              & \ms{2.10}{1.60}   & \ms{2.95}{1.10}       & \ms{1.83}{1.39}   & \textbf{\ms{1.25}{0.74}}     & \ms{2.21}{1.52}             & \ms{2.55}{1.34}    & \underline{\ms{1.56}{2.21}}      \\
\bottomrule
\end{tabular}}
    \end{minipage}  
\vspace{-1em}
\end{table*}

\noindent\textbf{Protocols.}
For the implementation of {\method}, we utilize a two-layer GraphSAGE~\citep{sage} as the feature embedding and inference mechanism. For {\vgae} and {\nodetovec}, we adhere to the hyperparameter settings outlined in~\citep{masrour2020bursting}, while for the other baselines, we follow the configurations provided in their respective original papers. To fine-tune the model, we perform a grid search over the hyperparameters $\alpha$, $\beta$, and $\gamma$ for each dataset. Specifically, $\alpha$ and $\beta$ are selected from the set $\{0.001, 0.005, 0.01, 0.05, 0.1, 0.5, 1.0, 1.5, 2, 2.5\}$, and $\gamma$ is chosen from $\{0.2, 0.4, 0.6, 0.8, 1.0, 1.2, 1.4, 1.6, 1.8, 2.0, 2.5\}$. Each experiment is conducted $10$ times, with training set to $200$ epochs across all datasets. The learning rate is specifically tuned for each dataset: $0.005$ for {\pubmed}, and $0.01$ for {\dblp}, {\google}, and {\fb}. All losses are optimized using the Adam optimizer~\citep{kingma2014adam}.

\subsection{Link Prediction and Fairness Performance of {\method}}
To evaluate the performance of our proposed method in both link prediction and fairness, we conducted a comprehensive comparison with the previously mentioned baselines using four real-world datasets. The results, which include the mean and standard deviations for all models across these datasets, are detailed in~\cref{tab:main_exp}. From these results, we can draw the following observations:
\begin{itemize}
\item Our proposed method, \method, consistently demonstrates superior fairness performance in terms of both $\Delta_{\mathit{DP}}$ and $\Delta_{\mathit{EO}}$ across all evaluated datasets. For example, compared to \vgae, \method reduces $\Delta_{\mathit{DP}}$ by $74.0\%$, $82.2\%$, $77.9\%$, and $59.1\%$ on the \pubmed, \dblp, \google, and \fb, respectively. 

\item Regarding utility, \method typically achieves the second-best performance in terms of F1-score and AUC. For instance, \method retains $97.08\%$, $99.26\%$, $96.61\%$, and $99.95\%$ of the F1-score of \vgae on the \pubmed, \dblp, \google, and \fb datasets, respectively.

\item Fair link prediction baselines, such as {\fairpage}, {\fairwalk}, {\fairadj}, {\flip}, and {\fairegm}, exhibit less predictive bias compared to standard link prediction models like {\vgae} and {\nodetovec}. Among these, \texttt{fairadj} generally performs second-best after {\method}. Specifically, {\fairadj} shows better performance on the {\fb}, while {\flip} outperforms the others on the {\google}.
\end{itemize}
\subsection{Fairness-utility Trade-off Comparison}
\begin{figure*}[!t]
\centering
\includegraphics[width=1.0\linewidth]{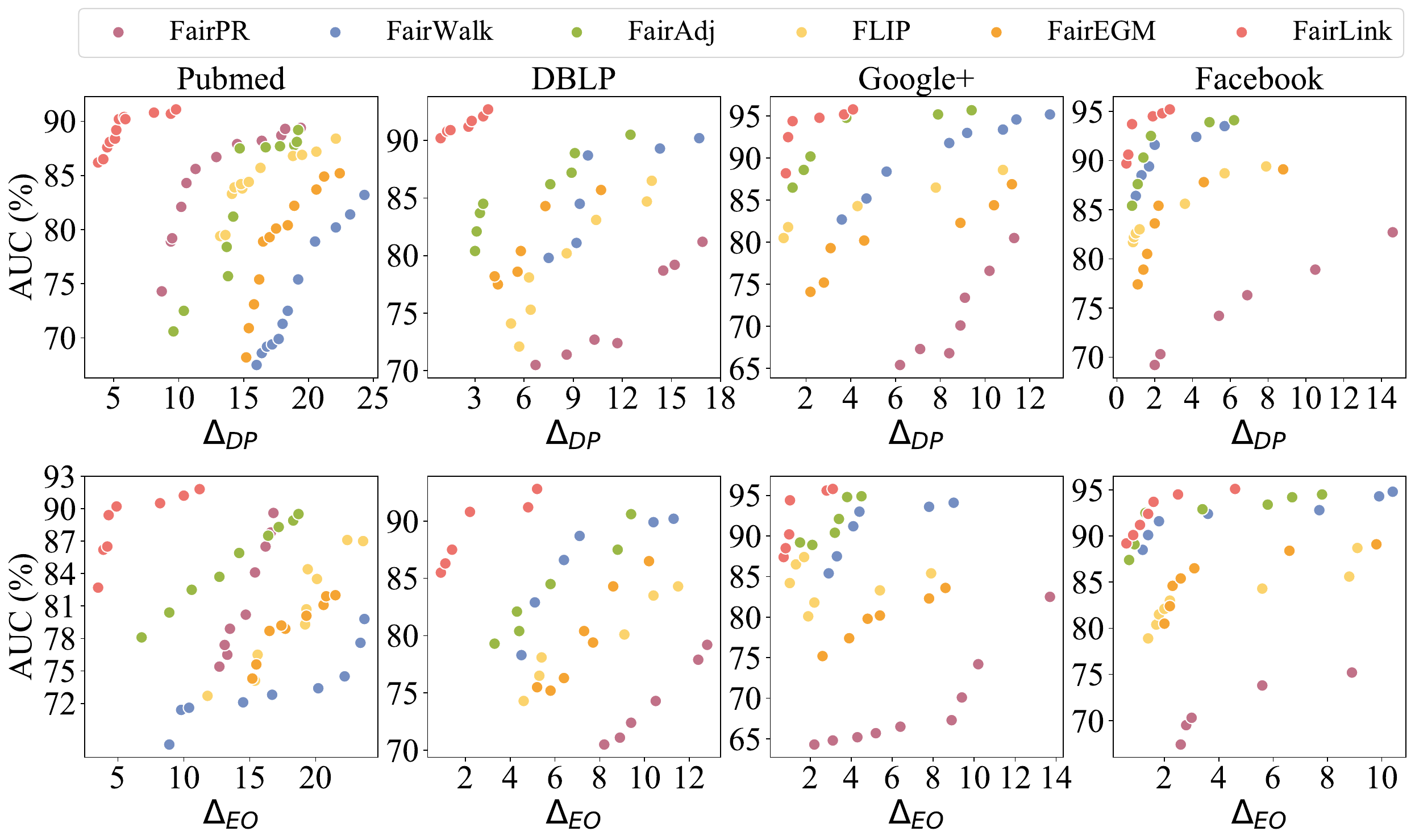}
\caption{Trade-off between fairness and link prediction accuracy across four datasets. Results in the upper left corner, which exhibit both lower bias and higher accuracy, represent the ideal balance.} \label{fig:tradeoff}
\end{figure*}
In~\cref{fig:tradeoff}, different colors are employed to distinguish between {\fairpage}, {\fairwalk}, {\fairadj}, {\flip}, {\fairegm}, and {\method}. Ideally, a debiasing method should be positioned in the upper-left corner of the plot to achieve the optimal balance between utility and unbiasedness. As depicted in the figures, methods based on {\method} generally provide the most favorable trade-offs between these two competing objectives. In contrast, while {\fairadj} usually offers superior debiasing with minimal utility loss, {\fairwalk} excels in maintaining high utility but is less effective in reducing bias. Although {\fairpage} can achieve reasonable unbiasedness in embeddings, it significantly compromises utility compared to {\method}, as illustrated in the {\dblp} and {\google} datasets. In contrast, {\fairegm} does not show a notable debiasing effect.

\subsection{Generalizability to Other Link Prediction Models}
\setcounter{figure}{4}
\setcounter{subfigure}{0}
\begin{subfigure}
\setcounter{figure}{4}
\setcounter{subfigure}{0}
    \centering
    \begin{minipage}[b]{0.5\textwidth}
         \centering
        {\includegraphics[height=0.41\linewidth]{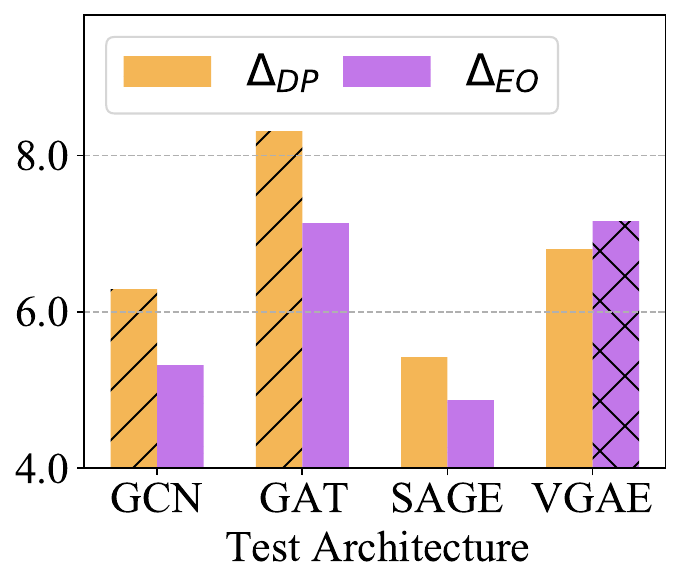}}\hspace{-2mm} 
        {\includegraphics[height=0.41\linewidth]{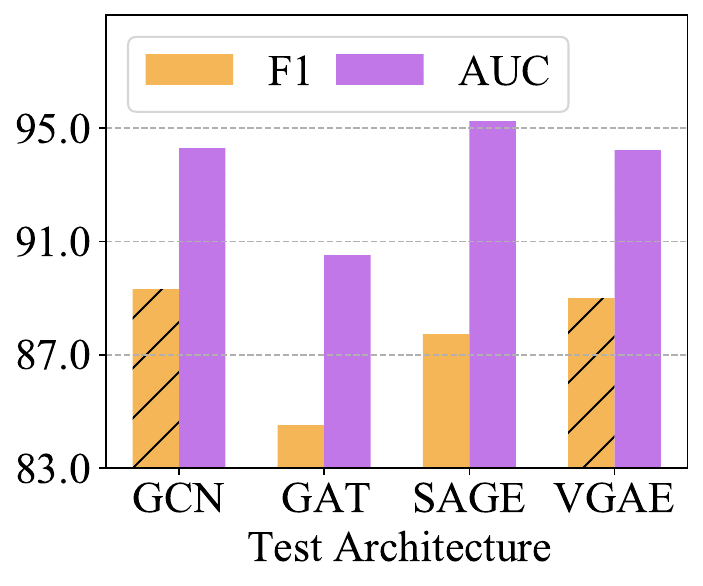} }
        \vspace{-0.5em}
        \caption{\pubmed.}
        \label{fig:cross_pubmed}
    \end{minipage}  
    \hspace{-2em}
\setcounter{figure}{4}
\setcounter{subfigure}{1}
    \begin{minipage}[b]{0.5\textwidth}
        {\includegraphics[height=0.41\linewidth]{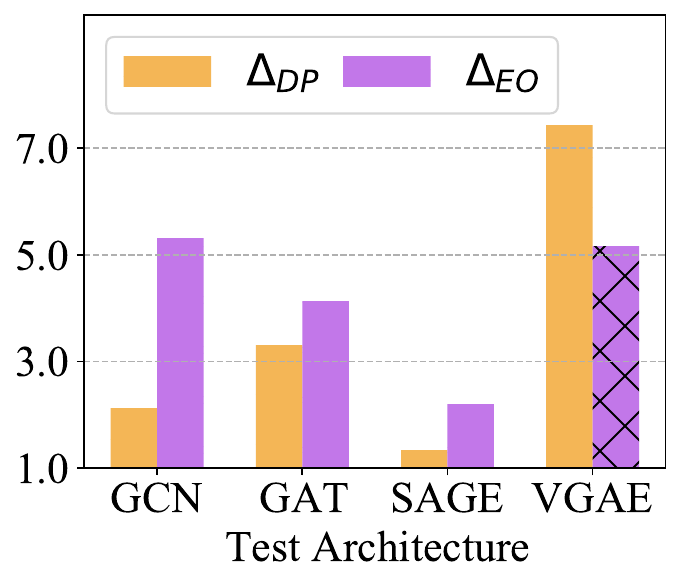} } \hspace{-2mm}
        {\includegraphics[height=0.41\linewidth]{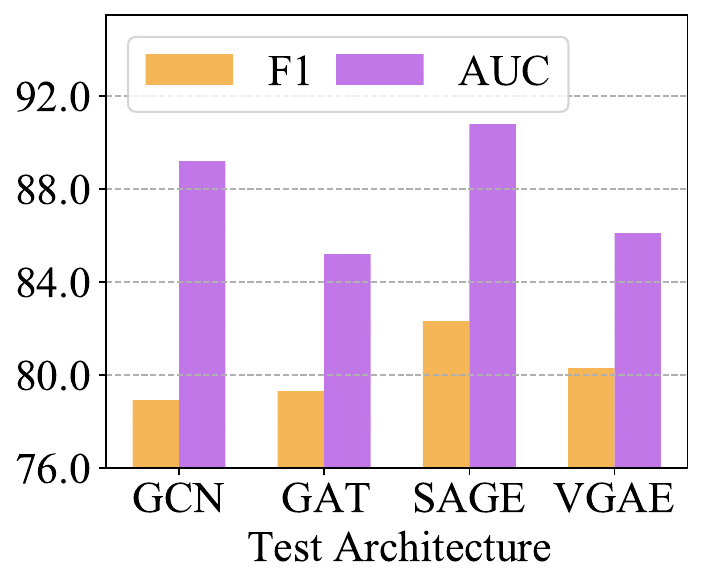} }
                \vspace{-0.5em}
        \caption{\dblp.}
        \label{fig:cross_dblp}
    \end{minipage}
    \\
\setcounter{figure}{4}
\setcounter{subfigure}{4}
    \begin{minipage}[b]{0.5\textwidth}
        {\includegraphics[height=0.41\linewidth]{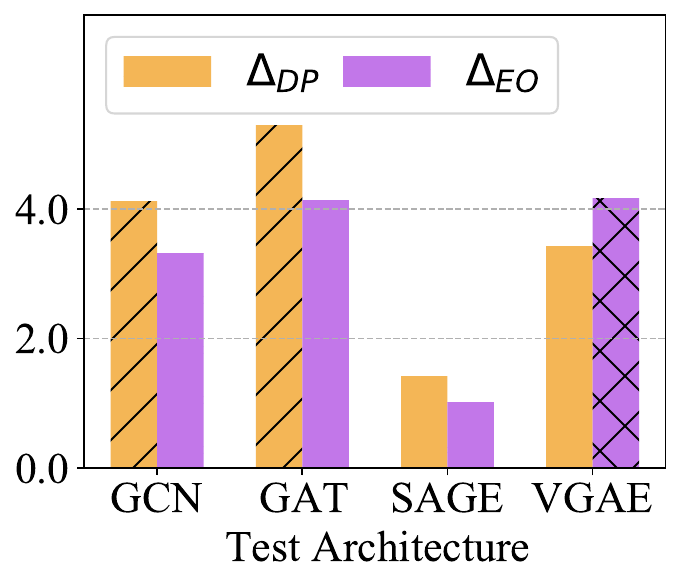} }\hspace{-2mm} {\includegraphics[height=0.41\linewidth]{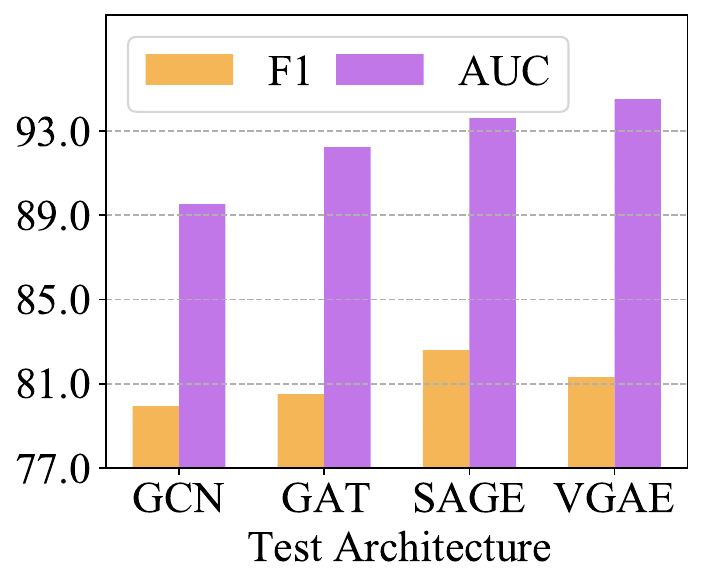} }
                \vspace{-0.5em}
        \caption{\google.}
        \label{fig:cross_google}
    \end{minipage}
   \hspace{-2em}
    \setcounter{figure}{4}
\setcounter{subfigure}{4}
    \begin{minipage}[b]{0.5\textwidth}
{\includegraphics[height=0.41\linewidth]{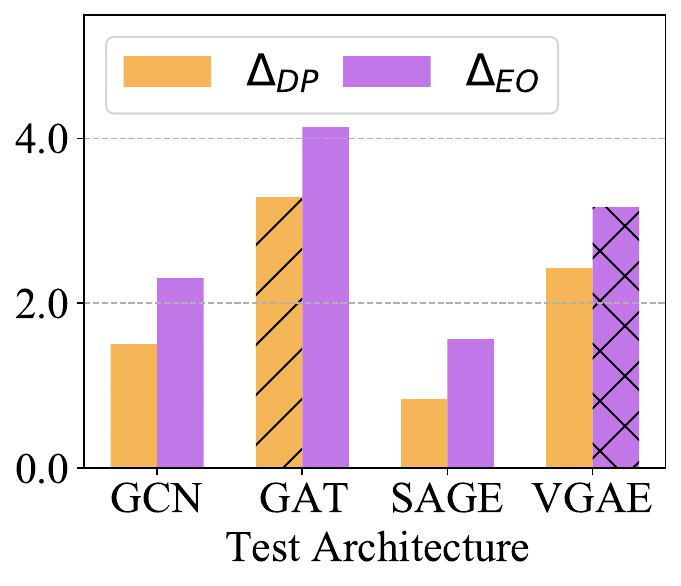} }\hspace{-2mm}{\includegraphics[height=0.41\linewidth]{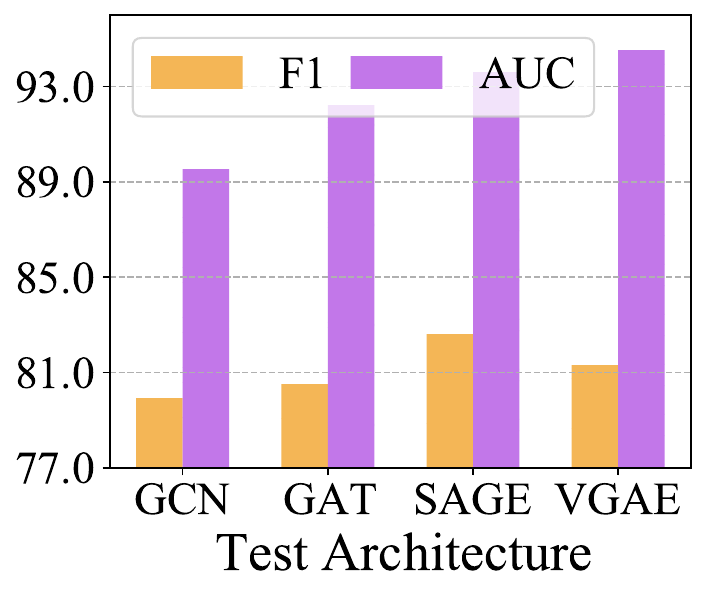} }
        \vspace{-0.5em}
        \caption{\fb.}
        \label{fig:cross_fb}
    \end{minipage}
\end{subfigure}
To validate the generalizability of the fairness-enhanced graph, we perform a cross-architecture analysis. Initially, we used GraphSAGE (SAGE) to generate synthetic graphs. These graphs are then evaluated across various architectures, including GCN, GAT, and VGAE, as well as on the original GraphSAGE model. Additionally, we apply {\method} with different structures to all datasets and assess the cross-architecture generalization performance of the fairness-enhanced graphs. The results of these experiments are documented in \cref{fig:cross_pubmed}, \cref{fig:cross_dblp}, \cref{fig:cross_google}, and \cref{fig:cross_fb}.

Compared to \cref{tab:main_exp}, \method demonstrates improved fairness performance over \vgae and \nodetovec across most model-dataset combinations. This indicates that \method is versatile and consistently achieves gains across various architectures and datasets. Our fairness-enhanced graphs show generally superior performance in fairness metrics (e.g., $\Delta_{\mathit{DP}}$ and $\Delta_{\mathit{EO}}$) and utility metrics (e.g., F1-score and AUC) across all datasets. Specifically, GraphSage excels in fairness across all datasets and achieves the best utility on \pubmed, \dblp, and \google.

\section{Related Work}
\subsection{Fairness in Machine Learning}
In recent years, numerous fairness definitions in machine learning have been proposed. These definitions generally fall into two categories: (1) \textit{Group fairness}, which aims to ensure that certain statistical measures are approximately equal across protected groups (e.g., racial or gender groups)~\citep{feldman2015certifying,hardt2016equality}; and (2) \textit{Individual fairness}~\citep{dwork2012fairness}, which does not rely on sensitive attributes but rather on the similarity between individuals. In our experiments, we adopt two widely used definitions of group fairness: demographic parity and equal opportunity. Demographic parity~\citep{feldman2015certifying} requires that members of different protected classes are represented in the positive class at the same rate, meaning the distribution of protected attributes in the positive class should reflect the overall population distribution. In contrast, equal opportunity~\citep{hardt2016equality} focuses on the model’s performance rather than just the outcome; it requires that true positive rates are equal across different protected groups, ensuring that the model performs consistently for all groups. Methodologically, existing bias mitigation techniques in machine learning can be broadly categorized into three approaches: (1) \textit{Pre-processing}, where bias is mitigated at the data level before training begins~\citep{calders2009building,kamiran2009classifying,feldman2015certifying}; (2) \textit{In-processing}, where the machine learning model itself is modified by incorporating fairness constraints during training~\citep{zafar2017fairness,goh2016satisfying}; and (3) \textit{Post-processing}, where the outcomes of a trained model are adjusted to achieve fairness across different groups~\citep{hardt2016equality}.

\subsection{Link Prediction}
Link prediction involves inferring new or previously unknown relationships within a network. It is a well-studied problem in network analysis, with various algorithms developed over the past two decades~\citep{liben2003link,al2006link,hasan2011survey}. Specifically, \textit{heuristic methods} define a score based on the graph structure to indicate the likelihood of a link's existence~\citep{liben2003link,newman2001clustering,zhou2009predicting}. The primary advantage of heuristic methods is their simplicity, and most of these approaches do not require any training. \textit{Graph embedding methods} learn low-dimensional node embeddings, which are then used to predict the likelihood of links between node pairs~\citep{grover2016node2vec,menon2011link}. These embeddings are typically trained to capture the structural properties of the graph. \textit{Deep neural network methods} have emerged as state-of-the-art for the link prediction task in recent years~\citep{gcn,sage,gat,kipf2016variational}. This category includes GNNs, which leverage the multi-hop graph structure through the message-passing paradigm. Additionally, GNNs augmented with auxiliary information, such as pairwise information~\citep{zhang2021labeling}, have been proposed to enhance link prediction. These advanced methods incorporate additional data to better capture the relationships between nodes~\citep{zhang2021labeling,zhu2021neural,wang2022equivariant}.

\subsection{Fair Link Prediction}\label{sec:related_fairlink}
With the success of GNNs, there has been increasing attention on ensuring fairness in graph representation learning~\citep{dai2022comprehensive}. Some works have focused on creating fair node embeddings, which are subsequently used in link prediction~\citep{bose2019compositional,buyl2020debayes,cui2018survey}. Others have directly targeted the task of fair link prediction~\citep{masrour2020bursting,li2021dyadic}. Specifically, \textit{dyadic fairness} has been proposed for link prediction, which requires the prediction results to be independent of whether the two vertices involved in a link share the same sensitive attribute~\citep{li2021dyadic}. To achieve dyadic fairness, the authors proposed {\fairadj}~\citep{li2021dyadic}, which leverages a variational graph auto-encoder~\citep{kipf2016variational} for learning the graph structure and incorporates a dyadic loss regularizer to enforce fairness. {FairPageRank (\fairpage)}~\citep{tsioutsiouliklis2021fairness} is a fairness-sensitive variation of the PageRank algorithm. It modifies the jump vector to ensure fairness, both globally and locally. The locally fair PageRank variant specifically guarantees that each node behaves in a fair manner individually. \textit{DeBayes}~\citep{buyl2020debayes} adopts a Bayesian approach to model the structural properties of the graph, aiming to learn debiased embeddings using biased prior conditional network embeddings. Meanwhile, {\fairwalk}~\citep{rahman2019fairwalk} adapts {\nodetovec}~\citep{grover2016node2vec} to enhance fairness in node embeddings by adjusting the transition probabilities in random walks, weighing the neighborhood of each node based on their sensitive attributes. Finally, {\flip}~\citep{masrour2020bursting} tackles graph structural debiasing by reducing homophily (the tendency of similar nodes to connect) in the graph. It measures fairness by assessing the reduction in modularity, a measure of the strength of the division of a graph into modules. {\fairegm}~\citep{current2022fairegm}, a collection of three methods that emulate the effects of a variety of graph modifications for the purpose of improving graph fairness.

\section{Conclusion}
We study fairness in link prediction. Existing methods primarily focus on integrating debiasing techniques during training to learn unbiased graph embeddings. However, these methods complicate the training process, especially when applied to large-scale graphs. Additionally, they are model-specific, requiring a redesign of the debiasing approach whenever the model changes. To address these challenges, we propose a data-centric debiasing method, {\method}, which aims to enhance fairness in link prediction without modifying the training of large-scale graphs. {\method} optimizes both fairness and utility by learning a fairness-enhanced graph. It minimizes the difference between the training trajectory of the fairness-enhanced graph and the input graph, incorporating fairness loss in the training of the fairness-enhanced graph. Extensive experiments on benchmark datasets demonstrate the effectiveness of {\method}, as well as its ability to generalize across different GNN architectures.

\section{Acknowledgements}
This work was supported in part by the DARPA Young Faculty Award, the US Army Research Office, the National Science Foundation (NSF) under Grants \#2127780, \#2319198, \#2321840, \#2312517, and \#2235472, the Semiconductor Research Corporation (SRC), the Office of Naval Research through the Young Investigator Program Award, and Grants \#N00014-21-1-2225 and \#N00014-22-1-2067. Additionally, support was provided by the Air Force Office of Scientific Research under Award \#FA9550-22-1-0253, along with generous gifts from Xilinx and Cisco.

\bibliographystyle{arxiv_submission}
\bibliography{main}

\newpage

\end{document}